\documentclass[fleqn,10pt]{wlscirep}
\usepackage[utf8]{inputenc}
\usepackage[T1]{fontenc}
\usepackage{graphicx}
\usepackage{gensymb}
\usepackage{amsmath}

\title{Prediction of progressive lens performance from neural network simulations}

\author[1,2]{Alexander Leube}
\author[1]{Lukas Lang}
\author[2]{Gerhard Kelch}
\author[1,2,*]{Siegfried Wahl}
\affil[1]{Institute for Ophthalmic Research, University of Tuebingen, Tuebingen, 72076, Germany}
\affil[2]{Carl Zeiss Vision International GmbH, Aalen, 73430, Germany}

\affil[*]{siegfried.wahl@uni-tuebingen.de}


\begin{abstract}
Purpose: Presbyopia, with its gradual loss of accommodation amplitude, is part of the natural
human aging process and commonly corrected using progressive addition lenses (PALs). However, the pre-clinical simulation of visual performance is difficult. Therefore, the purpose of this study is to present a framework to predict visual acuity (VA) based on a convolutional neural network (CNN) and to further to compare PAL designs.

Method: A simple two hidden layer CNN was trained to classify the gap orientations of Landolt Cs by combining the feature extraction abilities of a CNN with psychophysical staircase methods. The simulation was validated regarding its predictability of clinical VA from induced spherical defocus (between ±1.5 D,step size: 0.5 D) from 39 subjectively measured eyes. Afterwards, a simulation for a presbyopic eye corrected by either a generic hard or a soft PAL design (addition power: 2.5 D) was performed including lower and higher order aberrations.

Result: The validation revealed consistent offset of +0.20\,logMAR $\pm$0.035\,logMAR from simulated VA. Bland-Altman analysis from offset-corrected results showed limits of agreement ($\pm$1,96\,SD) of -0.08\,logMAR and +0.07\,logMAR, which is comparable to clinical repeatability of VA assessment. The application of the simulation for PALs confirmed a bigger far zone for generic hard design, but did not reveal zone width differences for the intermediate or near zone. Furthermore, a horizontal area of better VA at the mid of the PAL was found, which confirms the importance for realistic performance simulations using object-based aberration and physiological performance measures as VA.

Conclusion: The proposed holistic simulation tool was shown to act as an accurate model for subjective visual performance. Further, the simulation’s application for PALs indicated its potential as an effective method to compare visual performance of different optical designs. Moreover, the simulation provides the basis to incorporate neural aspects of visual perception and thus simulate the VA including neural processing in future.
\end{abstract}
\begin{document}

\flushbottom
\maketitle
\thispagestyle{empty}

\section*{Introduction}
Visual acuity as the major measure for visual performance in clinical and scientific studies is defined as the smallest, just resolvable detail in visual perception. The assessment of visual acuity follows standardized psychophysical procedures and is implemented in different clinical charts \cite{lim2010,ferris1982,sloan1959new}, computer-based tests \cite{rosser2003, bach2006} or mobile digital device applications on smartphones and tablets \cite{brady2015,tofigh2015,dubuisson2017}. The reduction of visual acuity, for instance from uncorrected refractive errors in far and near distance, leads to significant impairment in daily life tasks\cite{Lu2011}. Especially in the growing aged society optical corrections for near vision loss due to presbyopia become more and more important. Presbyopia, describing the condition of problems with focusing to near targets, is the most common occurring change of the aging eye \cite{patel2007} and its unmet need for optical correction lead to visual impairment, which is one of the greatest burden in rural areas of low-resource countries \cite{fricke2018}. Progressive addition lenses (PALs) are spectacle corrective lenses, which are characterized by a continuous change of optical power over their surface, which provide clear vision from far over intermediate to near objects \cite{sheedy2004a}. Such lenses are further suffering from unwanted peripheral astigmatic errors, which limits clear vision over the lower diagonals of the lens \cite{sheedy1987,minkwitz1963} and requires specific, situation-dependent eye-head coordination \cite{rifai2016}. The optical design of PALs, there mainly the spherical and astigmatic power distribution of the two lens surfaces are optimized, aims to widen the zones of high acuity and avoid adaptation issues. Especially in the development of new and individualized optical corrections for presbyopia like PALs \cite{gwiazda2003,solaz2008,chu2010,villegas2006} but also intra-ocular lenses \cite{deVries2013,findl2007} and multifocal contact lenses \cite{back1992,rajagopalan2006}, the testing of visual performance becomes essential, but is a long lasting procedure for the patients.

Therefore, previous studies made attempts to predict visual acuity under different optical conditions using mainly metric calculations and/or process models. With the advances of clincal wavefront aberrometry \cite{cheng2004}, the usage of Zernike description \cite{thibos2002} and the understanding of their interaction with the human visual performance \cite{applegate2000,applegate2003a,applegate2003b} increased. Thibos et al. (2004) \cite{thibos2004} developed 31 image quality metrics which can accurately predict sphero-cylindrical refractive errors and some, like the visual Strehl metrics, which computes the neuronal weighted optical quality of an image, also the visual acuity \cite{marsack2004}. Additionally to analytically modelling the visual acuity, other authors used ray-tracing in combination with physiological weighting functions, like the Stiles-Crawford effect \cite{stiles1939}, and further apply blur thresholds to simulate the resolution limit \cite{greivenkamp1995,klonos1996}. Process models are discussed to be superior to metrics, since they try to simulate the mechanism in human visual processing. Watson et al. (2008)\cite{watson2008} developed a model for the prediction of letter acuity, in which they apply template matching to optically filtered images. The template matching is performed using minimum distance or cross-correlation between the filtered images and the available template letter. This method is advantageous, since no favorably chosen thresholds are required. Similar to the work from Watson et al., a model by Nestares et al. (2003)\cite{nestares2003} is independent from parameters that would need to be adjusted, since it applies a Bayesian pattern-recognition method which is robust to the presence of optical aberrations. Such methods reveal correlations between simulated and clinical measured acuity values of 0.86\cite{watson2008} and 0.81\cite{marsack2004}, however need a deep knowledge of adjusting model parameter and a careful setting between different model stages.

The purpose of the current study was to develop a model combining physical optics, psychophysical staircase methods and convolutional neuronal networks to an integrated framework for the prediction of visual acuity. This approach was further applied to aberrated images originating from eye plus progressive addition lens aberrations in order to show its feasibility for comparative studies of PAL designs.

\section*{Results}
\subsection*{Comparison to clinical assessed VA}
\begin{figure}
  \centering
  \includegraphics[width=\linewidth]{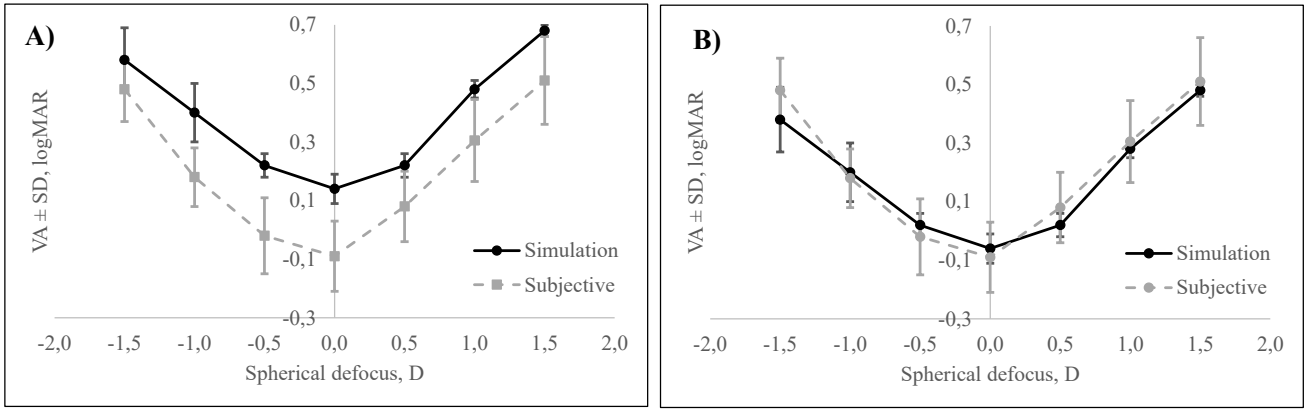}
  \caption{Mean VA $\pm$ inter-participants SD of 39 eyes for additional defocus between $\pm$1.5\,D. Simulated VA (red) is compared to VA from subjective measurements (blue)}
  \label{fig:Validation}
\end{figure}

Simulations from induced spherical defocus were validated to subjectively measured VA under comparable conditions. Fig. \ref{fig:Validation} shows the mean simulated vs. subjectively measured VA\,$\pm$\,standard deviation. The constant offset between subjectively measured and simulated VA for the different additional defocus values accounts +0.20\,logMAR ($\pm$0.035\,logMAR) throughout all spherical defocus steps. Figure \ref{fig:Validation}B provides the simulation plot corrected by the amount of the mean offset towards the subjective data plot. Further, for the subjective measurement, an asymmetry between negative and positive defocus is visible. After the offset correction, the simulated VA is slightly worse for negative defocus levels, while for positive defocus the simulation delivers marginally better VA results, compared to the subjectively measured VA. The standard deviations of the simulation for 1.5\,D and 1.0\,D are unexpectedly large compared to those for the other additional defocus values. The Bland-Altman analysis between the simulated and subjectively measured VA revealed the already mentioned bias of +0.20\,logMAR and further the consistency between the two datasets, as all data points are within the 95\% limit of agreement. After the correction of the offset the 95\% limit of agreement ranges from -0.08\,logMAR to +0.07\,logMAR.

\subsection*{Simulation of VA from PAL aberrations}
\begin{figure}
  \centering
  \includegraphics[width=0.8\linewidth]{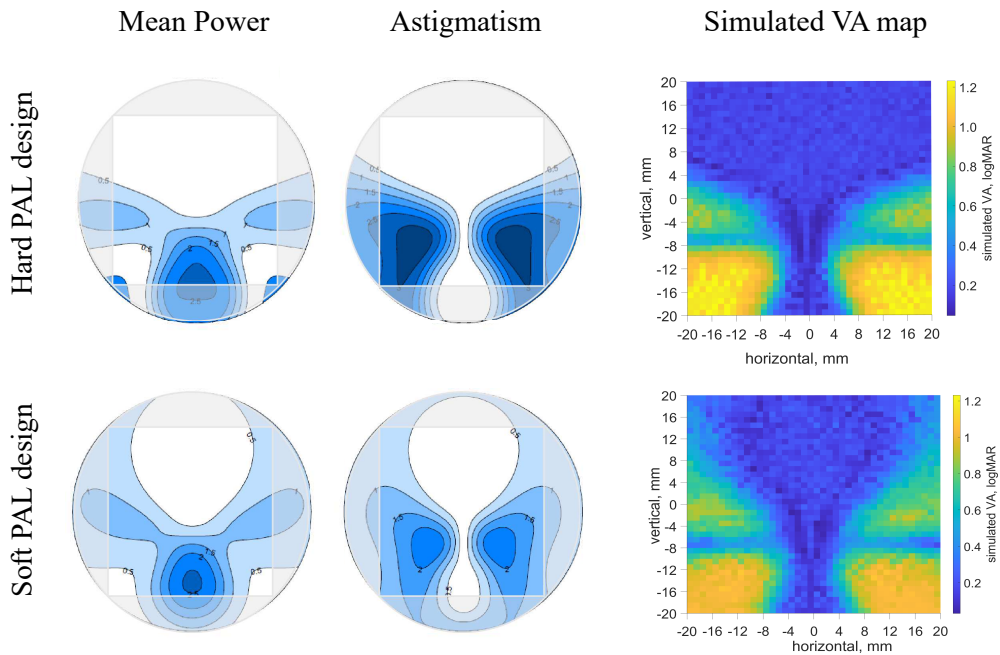}
  \caption{Simulated VA over the area of a hard (first row) and a soft (second row) PAL design. VA ranges from -0.2\,logMAR in blue to 1.1\,logMAR in yellow and is indicated for the central squared area of the PAL.}
  \label{fig:PAL_Results}
\end{figure}

\begin{figure}
  \centering
  \includegraphics[width=\linewidth]{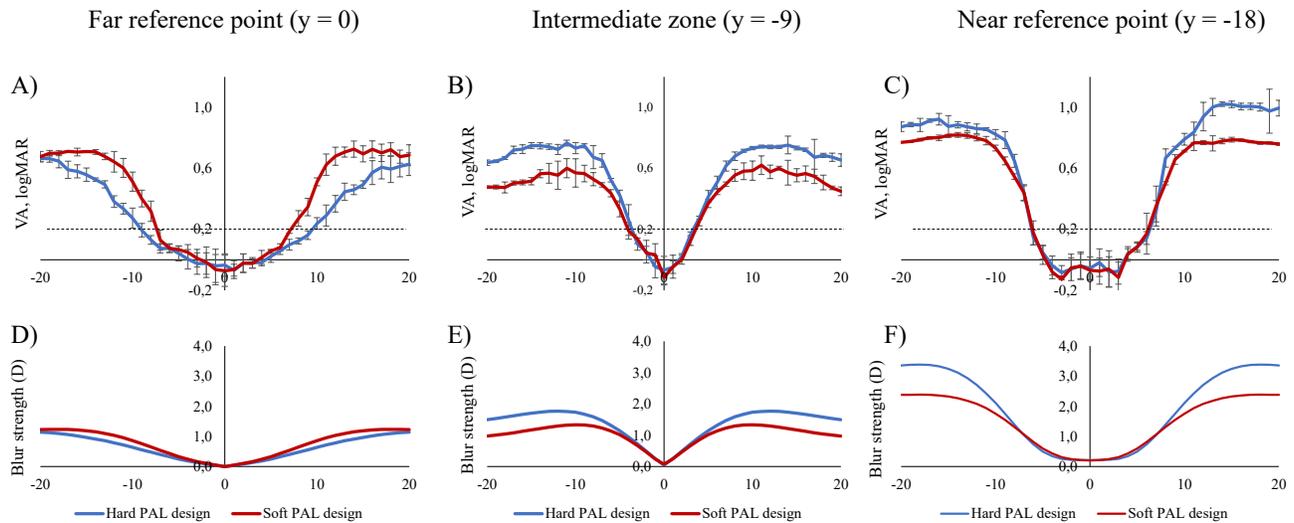}
  \caption{Three horizontal lines of simulated VA $\pm$ SD over the width of the hard (blue) and soft (red) PAL design in the upper row and comparison to the optical blur strength in the lower row. Visual acuity and blur strength are shown for the far reference point (y\,=\,0) in A and D, for the intermediate zone ( y\,=\,-9) in B and E and for the near reference point (y\,=\,-18) in C and F. Dashed line in the upper plots indicates a visual acuity level of 0.2\,logMAR.}
  \label{fig:PAL_Results_Zones}
\end{figure}

Fig. \ref{fig:PAL_Results} presents the results for the simulation of a generic hard and soft PAL design. The x- and y-axis represent the horizontal and vertical extension of the simulation area of the squared part from the PAL in millimeters. Moreover, the simulated VA is color-coded as a range from 0.0\,logMAR in blue to 1.2\,logMAR in yellow. The simulation’s offset of 0.20\,logMAR was corrected. The hard PAL design provides the better VA in the peripheral areas of the upper far zone of the lens. Both lens designs show a decreased VA at lower diagonals of the lens, as expected from the aberration map. These areas are interrupted by a striking continuous horizontal line of better VA, compared to the respective surrounding. In both lens designs, these areas are located at a vertical value of -7\,mm, from the far reference point and show a width of around 4\, mm. Furthermore, the simulation of the hard lens design provides an asymmetry between the VA in lower diagonals of the lens's simulated area.
For a more comprehensive comparison of the distant, intermediate and near zone from the generic hard and soft PAL design, the horizontal lines crossing the far reference point (y\,=\,0\,mm), the vertical center of the intermediate zone (y\,=\,-9\,mm) and the near reference point (y\,=\,-18\,mm) were investigated and are shown in Figure \ref{fig:PAL_Results_Zones}. The previously determined offset of the simulation was corrected. The widths of good VA ($\leq$\,0.2\,logMAR) for the three horizontal lines were compared and are indicated in the subfigures \ref{fig:PAL_Results_Zones} A-C as dashed lines. This width of good VA is 18\,mm for the hard and 14\,mm for the soft PAL design, 6\,mm and 7\,mm and 12\,mm and 12\,mm for the far, the intermediate and the near zone, respectively. 

\section*{Discussion}
\subsection*{Agreement between simulation and subjectively measured data}
The validation using a previously determined data set \cite{leube2018} revealed the simulation as a valid tool to model subjective VA measurements, based on wavefront aberrations. The simulated VA from the CNN stands up to the comparison of the participant’s VA in subjective testings. This was shown for different amounts of additional spherical defocus. The difference in offset between negative and positive spherical defocus is mainly based on the asymmetry of the subjective data, while the simulation led to only a marginal deviation between negative and positive spherical defocus. The asymmetry in the subjectively measured VA can be explained with a negative mean spherical equivalent refractive error from the cohort. Myopic participants show a lower sensitivity to negative induced defocus \cite{radhakrishnan2004a,radhakrishnan2004b} supposedly due to neural adaption to neural compensation of familiar aberrations \cite{artal2004}.

The main advantage of the method to simulate subjective VA, presented in the current study, compared to other models, for example by Guirao and Williams \cite{guirao2003} or Watson and Ahumada \cite{watson2008}, is not only that physical parameters like wavefront aberrations from several lens-eye combinations can be included, but also allow for agnosticly incorporated different stages of information of neural processing. This can be achieved by using training data for the CNN which includes the subjective response itself and therefore the neural interpretation by participants. However, this requires a large number of participants’ individual VA testings and therefore is left for further research. Generally, other CNNs used for object classification showed network architectures of a greater complexity than the CNN used in this simulation \cite{krizhevsky2012,ciresan2011}. However, it must be noted that the task of classifying Landolt C is rather simple, compared to visual interpretation tasks of natural surroundings. Therefore, the assumption was made and proved, that a very simple CNN architecture is sufficient to simulate subjective VA.

\subsection*{Application of the model as tool to compare PAL designs}
The simulated VA distributions from a generic hard and soft PAL design, allow a direct comparison of the two PAL designs towards physiological importance, since it's based on visual acuity. The main advantage of this simulation is the inclusion of higher order aberrations (HOAs) into the analysis of the PALs. Whereas, other methods are limited to   sphero-cylindrical errors (LOA) \cite{sheedy2004b,sheedy2004a}. Other approaches, which do provide the the ability of including the eye's aberrations \cite{villegas2006} are laboratory-based optical setups which can not provide fast results for a range of many different combinations.

A closer look at three selected horizontal lines of the PALs verified the suitability of the simulation as a tool to compare PAL designs. For the distant vision area and the intermediate area, the differences in the width of good VA of the hard and soft PAL designs represent the expected bigger far zone for the hard PAL designs. However, the similar size of the near zone width between the two PAL designs was unexpected, but could be due to the use of generic design which did not represent 'real' distribution of optical aberrations in PALs. Villegas et al. (2006) \cite{villegas2006} show that the eye's aberrations equalize the existing optical aberrations from the PAL, especially in the periphery, which could further explain the current finding. However, the simulated visual acuity distribution follows the calculated blur strength from the lens' LOAs, which was previously shown as a metric fairly predicting subjective visual performance \cite{marsack2004}. 

Obvious in both PAL designs, but more prominent in the soft design (see Figure \ref{fig:PAL_Results}), are little islands of better VA. They occur at the same vertical position of the PALs as the horizontal line in the simulation. Neither the astigmatism power, nor the astigmatism’s axes exhibit sudden transitions or show other striking features in that particular region. To investigate the possibility that the simulation causes these lines, the axes of the astigmatism and their possible interaction with certain orientations of the Landolt Cs gaps were taken into account. This follows the presumption that the axes of the astigmatism lead to asymmetries in the perceptual acuity \cite{wolffsohn2011,atchison2011,kobashi2012,leube2016}. Therefore, the simulation was performed while using only oblique orientated Landolt C and only straight oriented Cs. However, no differences were found. Qin et al. (2013) \cite{qin2013} stated in their study that the surface power of a PAL is not the same as the wearer power. They further show that an eye-lens optical system would provide more suitable measures for PAL optimization \cite{qin2013}. As also used in the current study, Wu et al. (2019) \cite{wu2019} described an object-based raytracing method to investigate optical aberrations in PALs. In their analysis using the modulation transfer at 10\,cycl/mm as performance metric, the simulated eye-lens system assuming a presbyopic eye corrected with a PAL showed similar horizontal 'islands' of good performance for the lens plane. In the current simulations, the presbyopia condition was mimicked as we normalized peripheral spherical errors to the central value along the umbilic line and therefore assume that the wearer uses always the correct optical zone in the PAL for the corresponding object distance. In combination with the equalization of aberrationsfrom the eye-lens systems, while using simulated visual acuity, this potentially led to the unusual appearance of the horizontal line of better VA. These findings would need to be confirmed by subjective measurements.

To conclude, it was shown that a simple CNN, combined with psychophysical methods, can be used for an accurate simulation of subjective visual performance. This approach provides the basis to modulate to holistic process of vision starting at the object side, considering the eye’s aberrations, creating the retinal image and finally including the neural processing of visual information, as potentially provided by further training data. We applied the simulation for the evaluation of PALs and showed the model as an effective instrument to evaluate the visual performance of different PAL designs. With the addition of neural processing, as mentioned above, this simulation will provide an alternative to long-lasting subjective measurements for the analysis of PAL designs. Moreover, it can be further developed towards a tool which determines the best-fitting PAL design for an individual, based on the individual wavefront aberrations of the eye.

\section*{Methods}
In the simulation of visual acuity (VA), a psychophysical staircase procedure and the classification abilities of a convolutional neural network (CNN) are combined to model subjective VA assessment. Therefore, the CNN is assumed to replace the human participant and to respond on the orientations of the presented standardized Landolt Cs. The input to the simulation is provided with the wavefront aberrations of an optical system in Zernike polynomials, which are used for calculating aberrated images. This approach is applied to aberration data from progressive addition lenses and is validated with clinically assessed acuity data.

\subsection*{Architecture and training of neuronal network}
The architecture of the simple CNN consists of nine layers, including two convolution layers and one fully-connected layer. 
All computations were performed in Matlab R2017a (The MathWorks Inc., Natick, USA). For its implementation and training the Neural Network Toolbox and Parallel Computing Toolbox from MathWorks were used.

The input image is provided in form of an artificially generated gray-scale image by the size of 512\,x\,512\,x\,1\,px. The first convolution layer applies 15 filters with a size of 5\,x\,5\,px to the pixel array of the input image. The stride was set to 1 and no padding was implemented. For the first convolution layer, this results in 15 feature maps of a size of 508\,x\,508 features which corresponds to a layer size of 508\,x\,508\,x\,15 features. In the second convolution layer the same filter, stride, and padding settings are repeated. Each convolution layer is followed by a ReLU. This double combination of convolution layers and ReLUs is succeeded by a max-pooling layer with a pool size of 2\,x\,2, a stride of 2, and no padding. The pooling leads to a reduction of the layer’s size. In the next step, the fully-connected layer creates the array of the output,
given by the eight categories of the Landolt Cs possible orientation. These orientations are: 0\degree, 45\degree, 90\degree, 135\degree, 180\degree, 225\degree, 270\degree and 315\degree. The array of the fully-connected layer is forwarded to the softmax layer which calculates a probability for each category. Finally the output layer presents the classification into the category of the highest probability.

The image data set to train the CNN contained 1\,762 images of Landolt rings per orientation.
From these, 1\,200 images were used for training and the remaining 562 images for testing. This leads to a total number of 14,096 images for training and 4\,496 images for testing. Each image of a Landolt C is labeled with its orientation. Images showing black high-contrast Landolt Cs on white background, with a gap size ranging between 3\,px and 80\,px, were used to train the network. This range of the ring’s sizes are related to a VA at 5\,m viewing distance between -0.19\,logMAR and 1.23\,logMAR. Since the network is supposed to be be trained on the prediction of VA under defocused conditions, additional spherical defocus in the range of $\pm$\,3.00\,D in steps of 0.25\,D was applied to these images \cite{legras2004} to decrease the quality of the images and train the CNNs abilities to classify blurry images. The Landolt Cs images had eight evenly distributed orientations, as mentioned above. The supervised training used a momentum of 0.9, a L2 regularization to avoid over-fitting and a mini-batch size of 128. The training was performed in parallel on the local GPU cores (NVIDIA GeForce GTX 1070, NVIDIA Corp., California, USA).


\subsection*{Psychophysical paradigm}
The previously trained network is embedded in a BestPEST staircase algorithm \cite{prins2016}, to simulate a clincal standard procedure for visual acuity testing \cite{bach2006}. The adaptive staircase approximates the visual acuity as a threshold for 50\% correct responses from a fitted logistic psychometric function. The slope of this function was fixed at $\beta\,=\,2.0$ and the stimulus range for the Landolt C gap size was set to $\alpha\,=\,[1:80]$ since it's limited by the total image size. The staircase included further a guessing rate of $\gamma\,=\,0.125$ and a lapse rate of $\lambda\,=\,0.02$. The aberrated image is presented to the previously trained CNN which classifies the image by the Landolt Cs gap orientation. The answer given by the network is compared to the actual orientation of the Landolt ring shown in the image. For correct or incorrect answers of the CNN, the gap size is decreased or increased, respectively, and the psychometric function is fitted to these responses. The staircase algorithm terminates after 30 trials and the gap size of the finally presented Landolt ring is used to calculate the MAR in arc minutes, there a pixel size in mm from a simulated monitor and a testing distance in mm between simulated observer and presented image is used. The VA is calculated as the logarithmic minimum angle of resolution. This complete procedure of the simulation is repeated 10 times to eliminate small fluctuations of the simulation. Finally, the VA is calculated as the mean of these repetitions.

\subsection*{Progressive addition lens simulations}
The input to the simulation is provided as the wavefront aberrations of an optical system, in particular of an eye - PAL system in Zernike polynomials \cite{thibos2002}. With these wavefront aberrations the point spread function (PSF) of the optical system is computed
for a wavelength of $\lambda\,=\,530\,nm$ and pupil radius of $r_{pupil}\,=\,4\,mm$. To show feasibility for comparing different PAL designs, the simulation was performed for one hard and one soft design. Both PALs provided an addition power of +2.5\,D. The data of the two PALs contained Zernike polynomials up to the $4^{th}$ order, calculated and indexed according to OSA/ANSI standard \cite{thibos2002}. These polynomials were calculated by a ray-tracing software, applying an object-based ray-tracing \cite{wu2019} included an adaption to the correct object distances for the different zones of the PALs. Further, the calculations were performed for the pupil plane, where the distance between the back vertex of the lens and the pupil plane, for straight gaze, was 16.5\,mm and the refractive index of the lens was n\,=\,1.530. The simulation was performed for a square area of 41\,x\,41\,mm of the lenses. We did not take into account the moving eye. The coordinate system to describe this area, ranges from -20\,mm to 20\,mm on the x- and y-axis for the horizontal and vertical expansion of the PALs, respectively. The simulation took place with a grid size of 1\,x\,1 mm, resulting in 1\,681 simulation points across the lens’ square area. The eyes aberrations are set to 0. To calculate the blur strength \cite{thibos2004} for the correct object distance of the different sections of the PAL, the mean spherical equivalent M of each point was normalized with reference to the central M (x = 0) of the corresponding to its respective vertical position.

\begin{figure}
  \centering
  \includegraphics[width=0.8\linewidth]{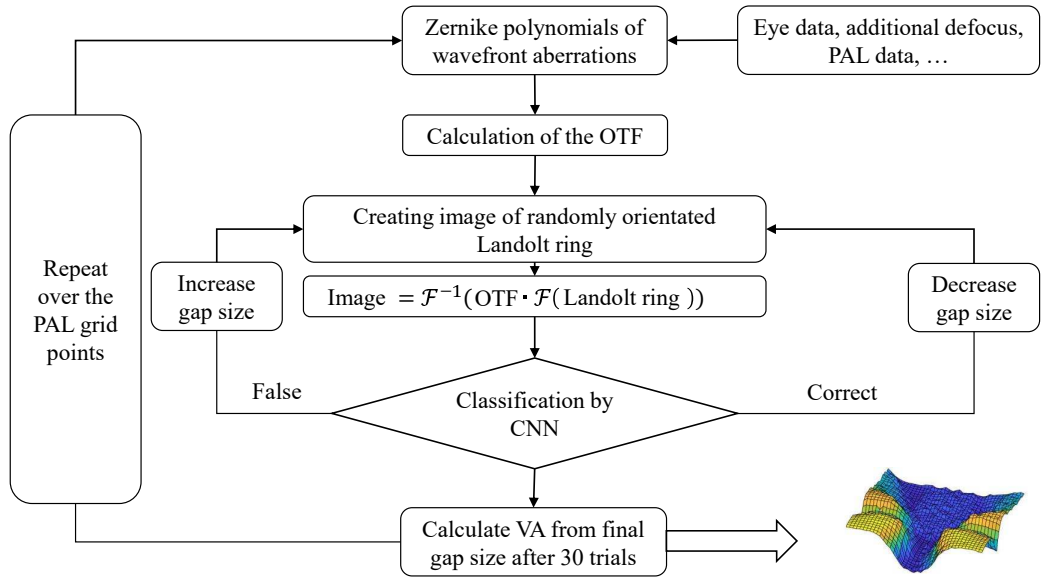}
  \caption{Schematic workflow for VA simulation combining physical optics, psychophysical staircase methods and convolutional neuronal network}
  \label{fig:Workflow}
\end{figure}

An image of a randomly orientated Landolt C is generated, prior to the simulation process. The Fourier transform of the PSF and the image are calculated separately and multiplied. An inverse Fourier transform is performed on the product to finally get the optically blurry image according to the optical system’s aberrations. This method was previously described by Legras et al. \cite{legras2004} and shown to provide correlated clinical measures \cite{leube2018,dehnert2011}. The workflow of the VA simulation of a PAL can be seen in Fig. \ref{fig:Workflow}.

\subsection*{Clinical validation of visual acuity simulation}
The validation of the simulation was carried out by the comparison of the simulated results with subjective data which was acquired under sphercial defocus. These data was provided from a former study \cite{leube2018} and includes wavefront aberration data of 39 eyes of participants in the age between 18 and 35 years. The aberrations of these eyes were measured with a commercially available wavefront aberrometer (i.Profiler plus, Carl Zeiss Vision GmbH, Germany). Further, the VA was evaluated using high-contrast Landolt rings at a range of induced spherical defocus of $\pm\,1.5\,D$ in steps of 0.5\,D. The defocus was induced by placing the corresponding lens in front of the eye for monochromatic ligth conditions. The consistency between the simulation and the subjectively measured data was evaluated with a Bland-Altman analysis. In general, the Bland-Altman analysis \cite{altman1983} illustrates the match between two different data sets. The limits of agreement were calculated as $\pm\,1.96$\,SD from the difference between the two methods.

\newpage
\bibliography{main}

\section*{Author contributions statement}

A.L., L.L. and S.W. conceived the experiment(s), A.L., L.L and G.K. conducted the simulations, A.L., L.L., G.K. and S.W. analysed the results. All authors reviewed the manuscript. 

\section*{Additional information}
This work was done in an industry-on-campus-cooperation between the University of Tubingen and Carl Zeiss Vision International GmbH. A.L., G.K. and S.W. are employed by Carl Zeiss Vision International GmbH and are scientists at the University of Tuebingen. There is no conflict of interest regarding this study. 
\end{document}